\documentclass[letterpaper, 10 pt, conference]{ieeeconf}  

\IEEEoverridecommandlockouts                              

\usepackage{cite}
\usepackage{amsmath,amssymb,amsfonts}
\usepackage{algorithmic}
\usepackage{graphicx}
\usepackage{textcomp}
\usepackage[dvipsnames]{xcolor}
\usepackage{multirow}
\usepackage{adjustbox}
\usepackage{caption}
\usepackage{subcaption}
\usepackage{balance}
\usepackage{float}
\usepackage{url}
\usepackage{hyperref}
\usepackage{multirow}
\usepackage{booktabs}
\usepackage{amssymb}
\usepackage{xcolor}

\newcommand{\namemethod}{V2V-GoT}
\newcommand{\namedataset}{V2V-GoT-QA}

\newcommand{\nameprior}{V2V-LLM}

\title{\LARGE \bf
\namemethod: Vehicle-to-Vehicle Cooperative Autonomous Driving with Multimodal Large Language Models and Graph-of-Thoughts
}

\author{
Hsu-kuang Chiu$^{1,2}$\quad Ryo Hachiuma$^1$\quad Chien-Yi Wang$^1$\quad  Yu-Chiang Frank Wang$^1$\quad \\ Min-Hung Chen$^{1*}$\quad Stephen F. Smith$^{2*}$\quad
\thanks{
$\textsuperscript{\rm 1}$NVIDIA,\quad$\textsuperscript{\rm 2}$Carnegie Mellon University,\quad$\textsuperscript{\rm *}$ equally advising.}
\thanks{The authors thank Boyi Li, Zhiding Yu, Boris Ivanovic, and Marco Pavone from NVIDIA for valuable discussions and feedback.}
\thanks{This research was funded by NVIDIA, the CMU Safety21 University Transportation Center, and CMU Robotics Institute.}
}

\begin{document}

\maketitle
\thispagestyle{empty}
\pagestyle{empty}

\begin{abstract}
Current state-of-the-art autonomous vehicles could face safety-critical situations when their local sensors are occluded by large nearby objects on the road. Vehicle-to-vehicle (V2V) cooperative autonomous driving has been proposed as a means of addressing this problem, and one recently introduced framework for cooperative autonomous driving has further adopted an approach that incorporates a Multimodal Large Language Model (MLLM) to integrate cooperative perception and planning processes.  However, despite the potential benefit of applying graph-of-thoughts reasoning to the MLLM, this idea has not been considered by previous cooperative autonomous driving research. In this paper, we propose a novel graph-of-thoughts framework specifically designed for MLLM-based cooperative autonomous driving. Our graph-of-thoughts includes our proposed novel ideas of occlusion-aware perception and planning-aware prediction. We curate the \namedataset~dataset and develop the \namemethod~model for training and testing the cooperative driving graph-of-thoughts. Our experimental results show that our method outperforms other baselines in cooperative perception, prediction, and planning tasks. Our code and dataset are released to facilitate open-source research at \href{https://eddyhkchiu.github.io/v2vgot.github.io/}{https://eddyhkchiu.github.io/v2vgot.github.io/}. 
\end{abstract}    
\section{Introduction}

Today's autonomous vehicles rely mainly on mounted cameras or LiDAR sensors to perceive the world, understand the dynamic surrounding scenes, and take driving decisions over time. Inherently such reliance on the vehicle's local sensors can be limiting, particularly in situations where vehicles and other potential obstacles are occluded by other large nearby objects, such as buses or trucks. To mitigate this safety issue, recent research has proposed mechanisms for V2V-enabled  cooperative perception~\cite{xu2022opencood, xu2022v2xvit, xu2022cobevt, chiu2023selective, chiu2024probabilistic,xiang2023hmvit,cui2022coopernaut,zhou2025turbo}. In a cooperative perception framework, each Connected Autonomous Vehicle (CAV) can share its individual perception information with others and improve overall detection and tracking accuracy. 

How to best realize cooperative perception remains a question for research. One recent
trend in autonomous driving research is using Large Language Models (LLMs) and their variants to build end-to-end driving systems that digest the raw sensor input and then generate the planning results. Such LLM-based autonomous driving models~\cite{sima2023drivelm, wang2025omnidrive, tian2024token} are motivated by the belief that LLM-based approaches could benefit from the reasoning and generalization capabilities that large-scale pre-trained data provides. However, most LLM-based driving research focuses on individual autonomous vehicles without cooperative perception or planning.

Some recent pioneering research works~\cite{chiu2025v2vllm, liu2025colmdriver, fang2025codrivingllm,gao2025langcoop} have started to explore LLM-based cooperative autonomous driving. However, the V2V cooperation in prior works~\cite{ liu2025colmdriver, fang2025codrivingllm} only involves the LLM as a pure language-based negotiator to resolve planning conflicts, which has not taken advantage of the multimodal understanding ability of MLLMs. In contrast, previous work \nameprior~\cite{chiu2025v2vllm} adopts MLLMs to fuse perception features from multiple CAVs and provide answers to perception and planning questions. Although \nameprior~\cite{chiu2025v2vllm} shows promising results, it has not considered the additional benefit of applying chain-of-thoughts or graph-of-thoughts reasoning on  MLLMs.

\begin{figure}[!t]
\centering
\includegraphics[width=0.47\textwidth]{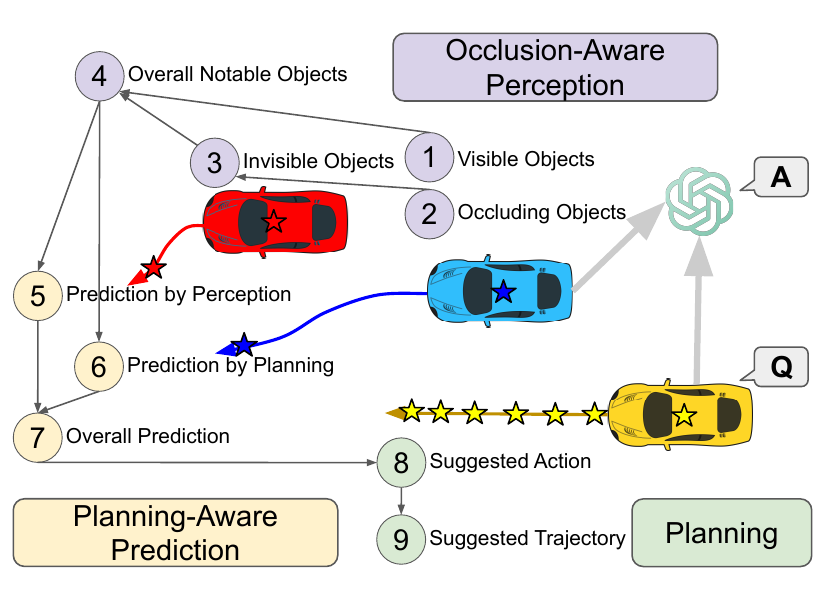}
\caption[]
        {Illustration of our proposed graph-of-thoughts reasoning framework for cooperative autonomous driving.
        All Connected Autonomous Vehicles (CAVs) share their perception features with the Multimodal Large Language Model (MLLM), as illustrated by the grey arrows. Any CAV can ask the MLLM to provide a suggested future trajectory or answer perception or prediction questions. The MLLM fuses the perception features from all CAVs and performs inference by following the graph-of-thoughts.
        If two QA nodes are connected by a directed edge in the graph, as illustrated by black arrows, the answer of the parent node QA is used as the input context of the child node QA. 
        Other colored curved arrows illustrate the predicted or suggested future trajectories.
        Color stars represent current locations of objects, predicted or suggested future waypoints.
        } 
        \vspace{-20pt}
        \label{fig:fig1}
\end{figure}

In this paper, we propose a new graph-of-thoughts reasoning framework designed for MLLM-based cooperative autonomous driving, as illustrated in Figure~\ref{fig:fig1}. We follow the problem setting from \nameprior~\cite{chiu2025v2vllm}, where all CAVs share their perception features with a  MLLM. Any CAV can ask the MLLM to answer a driving question. The MLLM fuses the perception features from all CAVs and performs inference to provide the answer. Unlike previous work~\cite{chiu2025v2vllm,sima2023drivelm}, we propose a novel graph-of-thoughts reasoning framework specifically designed for cooperative autonomous driving scenarios, as illustrated in Figure~\ref{fig:fig1} and \ref{fig:all_q_illustration}.

To verify the effectiveness of our proposed idea, we first curate the  \namedataset~dataset, built upon the V2V4Real~\cite{xu2023v2v4real} cooperative driving dataset. Our \namedataset~includes $9$ different types of QA samples, as illustrated in Figure~\ref{fig:all_q_illustration}. Our proposed \textbf{occlusion-aware perception} questions (Q1 - Q4) consider visible, occluding, and invisible objects. Our proposed \textbf{planning-aware prediction} questions (Q5 - Q7) include prediction by perception features and prediction by other CAVs' current planned future trajectories. Our planning questions (Q8 - Q9) provide the suggested action settings and suggested waypoints of future trajectories to avoid potential collisions. Different types of QA are connected by a directed edge in our graph-of-thoughts, as shown in Figure~\ref{fig:fig1}. The answer of the parent node QA is used as the input context of the child node QA. 

In addition to the newly curated dataset, we also develop our baseline model \namemethod, as shown in Figure~\ref{fig:model}. Unlike the prior work \nameprior~\cite{chiu2025v2vllm} and DriveLM~\cite{sima2023drivelm} that only take the perception features at the current timestep as visual input, our \namemethod~includes the perception features from both current and previous timesteps to better capture the temporal dynamics of the surrounding objects.

In our experiments, we compare the final planning performance of our proposed \namemethod~and other baseline methods from the prior work \nameprior~\cite{chiu2025v2vllm}, including \textit{no fusion}, \textit{early fusion}, and \textit{intermediate fusion}. Experimental results show that our model outperforms all other baselines in the planning tasks, achieving the lowest collision rates and L2 errors. Furthermore, our ablation study also verifies that the newly proposed QA types in our graph-of-thoughts, namely occlusion-aware perception QAs (Q1 - Q4) and planning-aware prediction QAs (Q5 - Q7), are helpful in achieving better cooperative perception, prediction, and planning performance for MLLM-based cooperative autonomous driving.

Our contribution can be summarized as follows.
\begin{itemize}
\item  We present a novel graph-of-thoughts reasoning framework for MLLM-based cooperative autonomous driving. The approach includes QA types designed specifically for cooperative driving, including \textbf{occlusion-aware perception} and \textbf{planning-aware prediction}.

\item We curate the \namedataset~dataset and develop the \namemethod~model to establish the benchmark and baseline method for future comparative research on MLLM-based cooperative autonomous driving with graph-of-thoughts reasoning.

\item Experimental results demonstrate the effectiveness of \namemethod~in improving overall perception, prediction, and planning performance for cooperative autonomous driving.

\end{itemize}

\section{Related Work}

\subsection{Vehicle-to-Vehicle Cooperative Perception}
V2V cooperative perception algorithms are proposed to improve the overall detection and tracking accuracy in cooperative driving scenarios by sharing perception information between CAVs. Pioneering works F-cooper~\cite{chen2019fcooper} and V2VNet~\cite{wang2020v2vnet} propose the intermediate fusion approach that shares perception feature maps, achieving a good balance between performance and communication cost. V2V4Real~\cite{xu2023v2v4real} is the first real cooperative driving dataset available worldwide with  detection and tracking benchmarks that evaluate the real-world performance of numerous abundant cooperative detection~\cite{xu2022opencood,xu2022v2xvit,xu2022cobevt} and cooperative tracking algorithms~\cite{chiu2024probabilistic, su2024coop}. However, this group of research focuses on perception and has not explored the planning part of autonomous driving. In contrast, our work covers both cooperative perception and planning.

\subsection{LLM-based Autonomous Driving}
A more recent autonomous driving research trend is applying an LLM to build end-to-end autonomous driving models due to its promising reasoning ability. Pioneering work GPT-driver~\cite{mao2023gpt, mao2023agentdriver} encodes the state of the ego vehicle and its object detection results into text and allows LLM to identify notable objects and suggest a future trajectory. Other works~\cite{qian2023nuscenesqa,wu2023nuprompt} apply  MLLMs to perform scene understanding from the image input. More advanced work, DriveLM~\cite{sima2023drivelm}, and LingoQA~\cite{marcu2023lingoqa} develop MLLM-based end-to-end models to further generate the suggested action or future trajectory as output. DriveLM~\cite{sima2023drivelm} also considers graph-of-thoughts but only for a single autonomous vehicle without V2V cooperation. This group of research focuses only on individual autonomous vehicles without cooperative perception or planning. In contrast, our work develops MLLMs-based models for V2V cooperative driving.

\subsection{LLM-based Cooperative Autonomous Driving}
A few very recent works~\cite{chiu2025v2vllm,fang2025codrivingllm,liu2025colmdriver,gao2025langcoop} start exploring LLM-based cooperative autonomous driving. CoDrivingLLM~\cite{fang2025codrivingllm} uses an LLM as a pure language-based conflict coordinator in simulated cooperative driving scenarios in an intersection. CoLMDriver~\cite{liu2025colmdriver} uses a Vision Language Model (VLM) as the initial planner of an individual autonomous vehicle without cooperation, while the V2V cooperation module still uses an LLM as a negotiator based on pure language in simulated environments. Neither prior work has taken advantage of
the multimodal understanding ability of MLLMs in their V2V cooperation modules nor verified their methods in real-world cooperative driving datasets. Another pioneering work \nameprior~\cite{chiu2025v2vllm} proposes an MLLM-based cooperative driving model and shows promising results in real-world datasets. However, the chain-of-thoughts or graph-of-thoughts reasoning capabilities have not been explored in this prior work. In contrast, our work proposes a novel graph-of-thoughts reasoning framework, including the special QAs designed for cooperative driving, such as occlusion-aware perception and planning-aware prediction.

\begin{figure*}[!t]
        \centering
        \begin{subfigure}[t]{0.32\textwidth}
            \centering 
            \includegraphics[width=\textwidth]{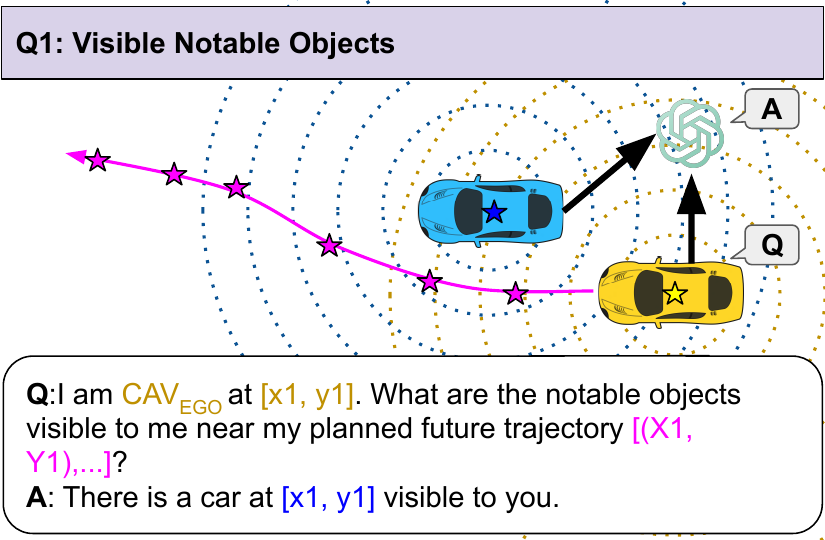}
            \caption[]%
            {{Q1: Visible Notable Objects.}}    
            \label{fig:q1_illustration}
        \end{subfigure}
        \hfill
        \begin{subfigure}[t]{0.32\textwidth}  
            \centering 
            \includegraphics[width=\textwidth]{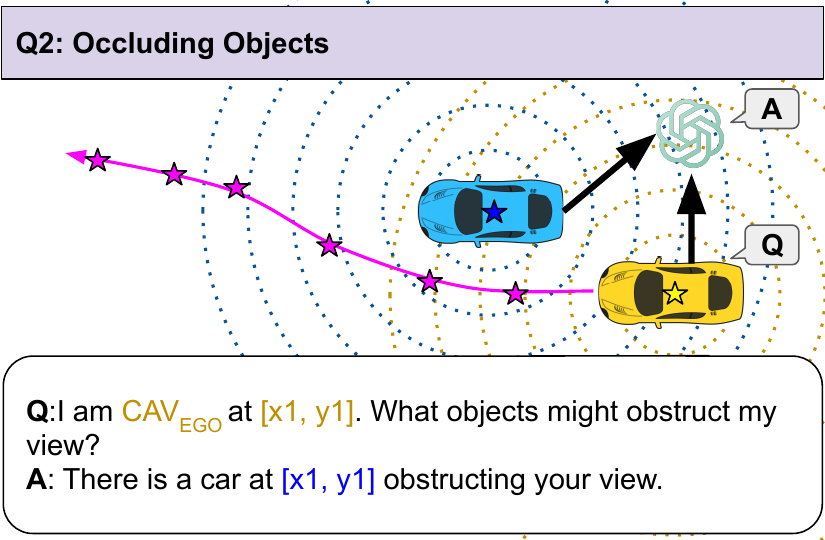}
            \caption[]%
            {{Q2: Occluding Objects.}}    
            \label{fig:q2_illustration}
        \end{subfigure}
        \hfill
        \begin{subfigure}[t]{0.32\textwidth}
            \centering 
            \includegraphics[width=\textwidth]{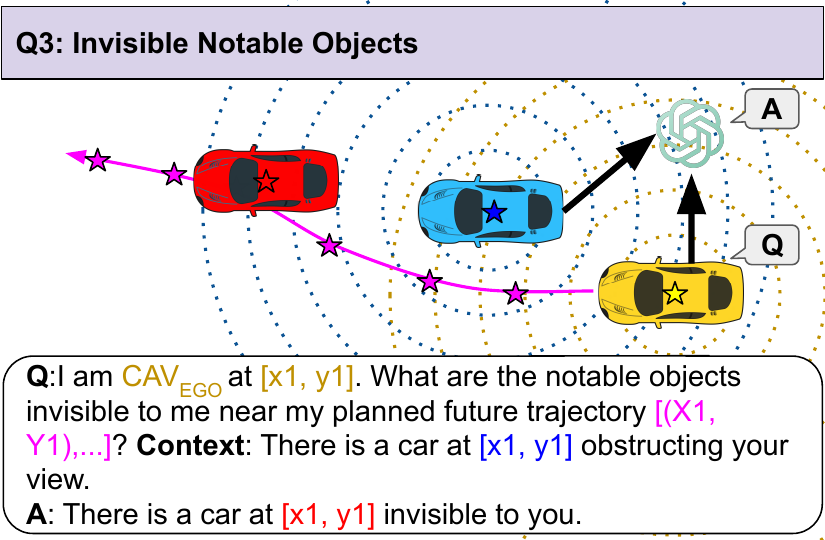}
            \caption[]%
            {{Q3: Invisible Notable Objects.}}    
            \label{fig:q3_illustration}
        \end{subfigure}

        \begin{subfigure}[t]{0.32\textwidth}
            \centering 
            \includegraphics[width=\textwidth]{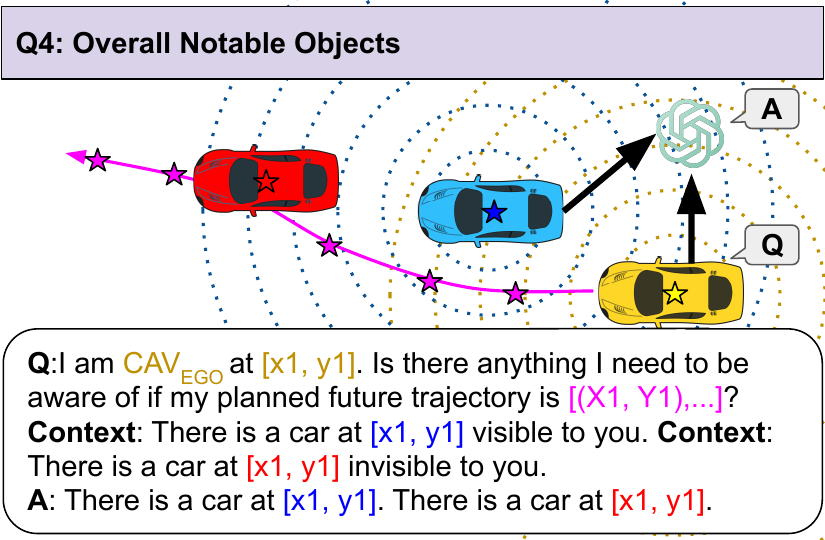}
            \caption[]%
            {{Q4: Overall Notable Objects.}}    
            \label{fig:q4_illustration}
        \end{subfigure}
        \hfill
        \begin{subfigure}[t]{0.32\textwidth}
            \centering 
            \includegraphics[width=\textwidth]{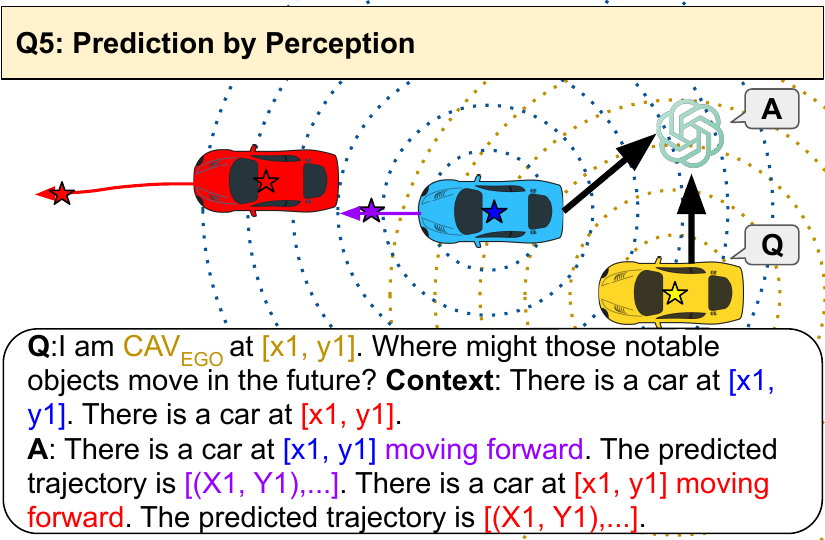}
            \caption[]%
            {{Q5: Prediction by Perception.}}    
            \label{fig:q5_illustration}
        \end{subfigure}
        \hfill
        \begin{subfigure}[t]{0.32\textwidth}
            \centering 
            \includegraphics[width=\textwidth]{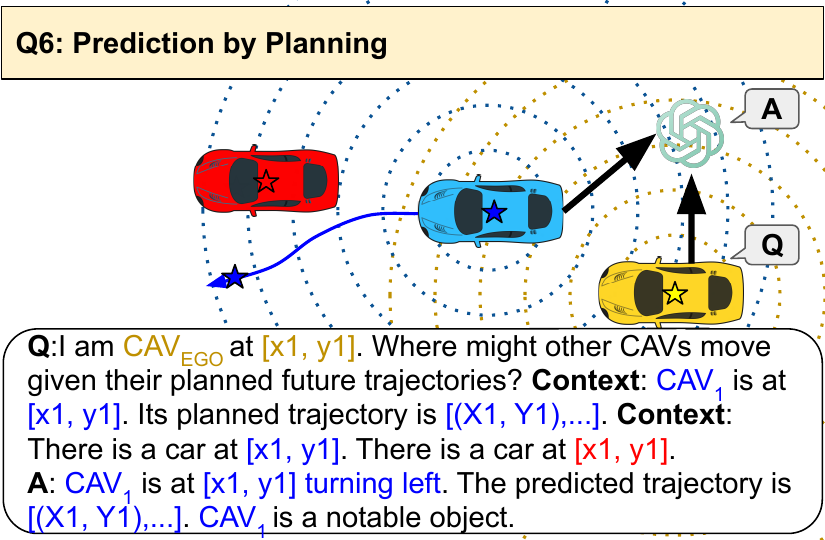}
            \caption[]%
            {{Q6: Prediction by Planning.}}    
            \label{fig:q6_illustration}
        \end{subfigure}

        \begin{subfigure}[t]{0.32\textwidth}
            \centering 
            \includegraphics[width=\textwidth]{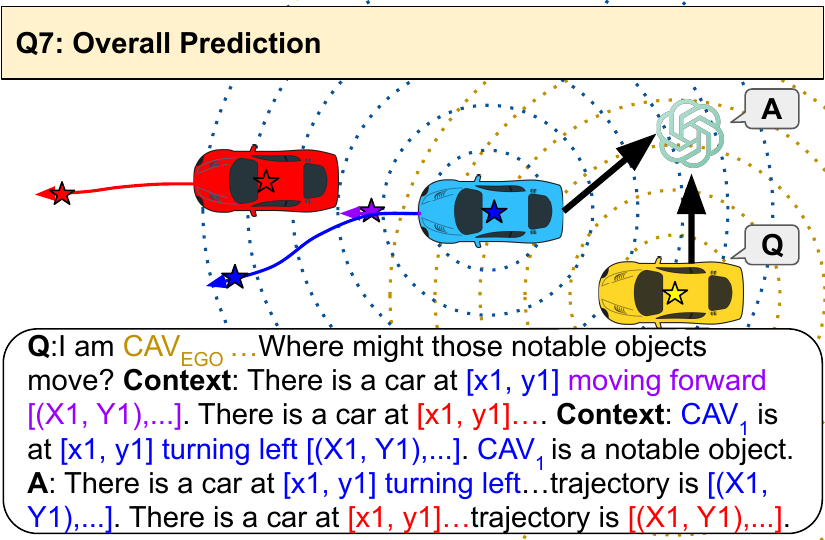}
            \caption[]%
            {{Q7: Overall Prediction.}}    
            \label{fig:q7_illustration}
        \end{subfigure}
        \hfill
        \begin{subfigure}[t]{0.32\textwidth}
            \centering 
            \includegraphics[width=\textwidth]{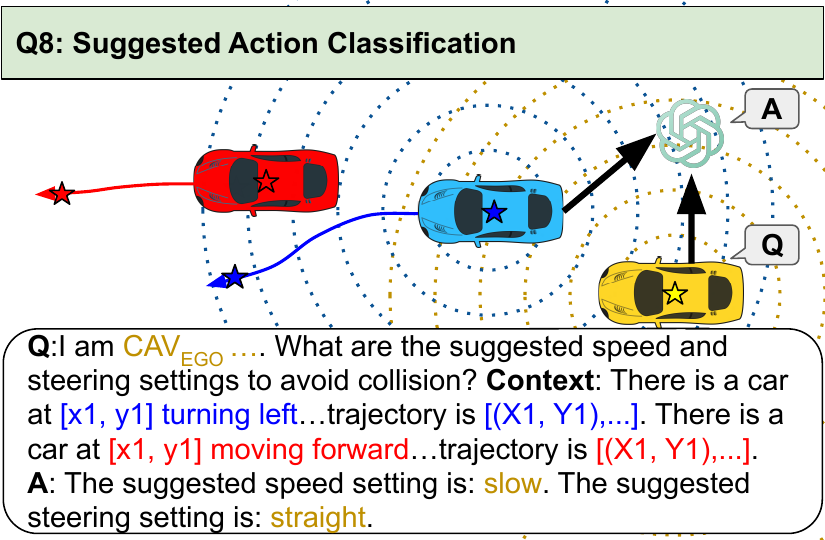}
            \caption[]%
            {{Q8: Suggested Action Classification.}}    
            \label{fig:q8_illustration}
        \end{subfigure}
        \hfill
        \begin{subfigure}[t]{0.32\textwidth}
            \centering 
            \includegraphics[width=\textwidth]{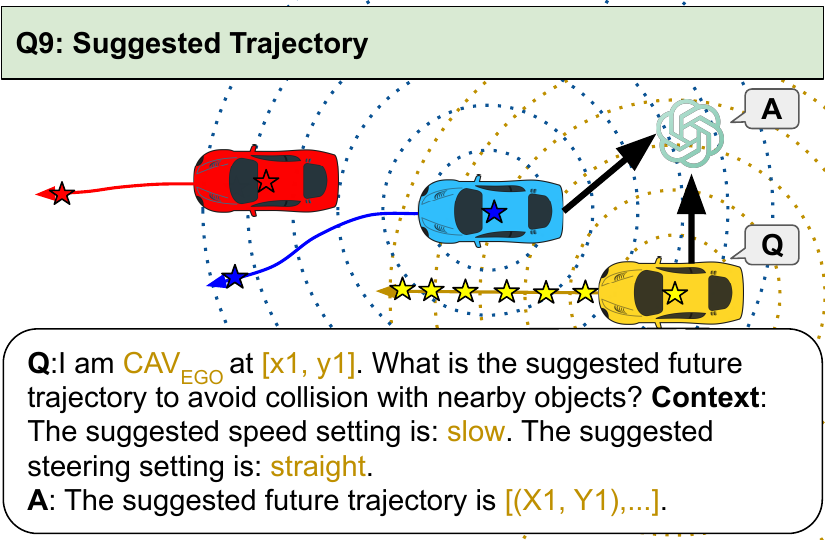}
            \caption[]%
            {{Q9: Suggested Trajectory.}}    
            \label{fig:q9_illustration}
        \end{subfigure}
           
        \caption[]
        {
        Illustration of \namedataset's $9$ types of QA pairs: Perception (Q1 - Q4), Prediction (Q5 - Q7), and Planning (Q8 - Q9). The black arrows pointing at the MLLM indicate the perception data from CAVs. Other colored arrows represent predicted or suggested future trajectories.
        } 
        \label{fig:all_q_illustration}
\end{figure*}

\section{\namedataset~Dataset}

\subsection{Problem Setting}
We build on the MLLM-based cooperative autonomous driving problem setting from V2V-LLM~\cite{chiu2025v2vllm}. The MLLM fuses the perception features from all CAVs and performs inference to answer a question. Unlike the prior work V2V-LLM~\cite{chiu2025v2vllm}, we propose a novel graph-of-thoughts reasoning framework, including the ideas of \textbf{occlusion-aware perception} and \textbf{planning-aware prediction}. We design $9$ types of question-answer pairs in our  \namedataset~dataset, including perception, prediction, and planning, as illustrated in Figure~\ref{fig:all_q_illustration}. Different types of QAs are connected by directed edges in the graph-of-thoughts, as shown in Figure~\ref{fig:fig1} and ~\ref{fig:full_graph}. The answer of the parent node QA is used as the input context of the child node QA.

\subsection{Dataset Details}
Following V2V-LLM~\cite{chiu2025v2vllm}, we use V2V4Real~\cite{xu2023v2v4real} as the base dataset and curate various QA pairs. V2V4Real~\cite{xu2023v2v4real} has $7105$ training frames and $1993$ testing frames. For each frame that is 3 seconds before the end of each driving sequence in V2V4Real~\cite{xu2023v2v4real}, we create one QA pair for each CAV per QA type. Overall, the resulting \namedataset~dataset has $110610$ training and $31014$ testing QA pairs.

\subsection{Dataset Curation}
We use the ground-truth bounding box annotations, each CAV and object's future trajectory, and their geometric relationships in V2V4Real~\cite{xu2023v2v4real} to curate the QA pairs of \namedataset~dataset. In our proposed graph-of-thoughts structure, the answer of the parent node QA is used as the input context of the child node QA. For training data, we use the ground-truth answer to generate the input context of the child node QA. During inference, we use the model inference output of the parent node QA to generate the input context of the child node QA. 

The QA types included in the \namedataset~dataset span tasks relating to Perception (Q1 - Q4), Prediction (Q5 - Q7), and Planning (Q8 - Q9). Details of each QA design rationale, how they are connected, dataset curation, and evaluation metrics are described in the following.

\subsection{Perception QAs}
The main goal of the perception QAs is to identify notable objects near ego CAV's current planned future trajectory. We propose the idea of \textbf{occlusion-aware perception} to the graph-of-thoughts reasoning and divide this task into subtasks of identifying notable objects that are visible (Figure~\ref{fig:q1_illustration}) and invisible (Figure~\ref{fig:q3_illustration}) to the ego CAV separately and then merge the results. This divide-and-conquer design can potentially make it easier for the model to learn which part of the perception feature maps to focus on in order to find the notable objects.

\noindent\textbf{Q1. Visible Notable Objects (Figure~\ref{fig:q1_illustration}):} We ask the MLLM to identify notable objects that are visible to the ego CAV and near its current planned future trajectory. Six waypoints from a planned future trajectory in the next 3 seconds are used as the reference trajectory in the question. To generate the ground-truth answer in the training data, at most 3 ground-truth objects within 10 meters of the reference trajectory that are detected by the ego CAV's individual 3D object detector are used as the ground-truth answer. We use the F1 score as the evaluation metric.

\noindent\textbf{Q2. Occluding Objects (Figure~\ref{fig:q2_illustration}):} Before identifying notable objects invisible to the ego CAV, locating the occluding objects first may provide useful information. To generate the ground-truth answer in the training data, at most 3 ground-truth objects closest to the ego CAV that are not occluded by other objects are used as the answer. We use the F1 score as the evaluation metric.

\noindent\textbf{Q3. Invisible Notable Objects (Figure~\ref{fig:q3_illustration}):} We ask the MLLM to identify notable objects that are invisible to the ego CAV and near its current planned future trajectory. In addition to the question, we also provide the input context from the answer of Q2. Occluding Objects (Figure~\ref{fig:q2_illustration}). Such context could be useful for the model to identify notable objects that are occluded from the ego CAV's field of view by focusing on the occluded region but in other CAVs' perception feature maps. To generate the ground-truth answer in the training data, at most 3 ground-truth objects within 10 meters of the reference trajectory that are not detected by the ego CAV's individual 3D object detector are used as the answer. We use the F1 score as the evaluation metric.

\noindent\textbf{Q4. Overall Notable Objects (Figure~\ref{fig:q4_illustration}):} This question simply takes the answers from Q1. Visible Notable Objects (Figure~\ref{fig:q1_illustration}) and Q3. Invisible Notable Objects (Figure~\ref{fig:q3_illustration}) as the input context and merges them to generate the final notable object identification answer, which are then used as the input context for the subsequent prediction questions. We use the F1 score as the evaluation metric.

\subsection{Prediction QAs}
Predicting the future trajectories of notable objects is critical for the final planning task to suggest a future trajectory for the ego CAV that can avoid potential collisions. We propose the idea of \textbf{planning-aware prediction} to the graph-of-thoughts reasoning and divide the prediction task into two parts: prediction by perception (Figure~\ref{fig:q5_illustration}) and prediction by planning (Figure~\ref{fig:q6_illustration}), and then merge their results to generate the final prediction output.

\noindent\textbf{Q5. Prediction by Perception (Figure~\ref{fig:q5_illustration}):} We ask the MLLM to predict the future trajectories of the notable objects and classify their movement into one of the following 4 categories: \textit{moving forward}, \textit{turning left}, \textit{turning right}, and \textit{staying at the same location}. In addition to the question, the MLLM also takes the answer of Q4. Overall Notable Objects (Figure~\ref{fig:q4_illustration}) as context. With this context, the MLLM knows where the notable objects are at the current timestep. This prediction process relies mainly on the perception features in the current and previous timesteps from all CAVs. We use the ground-truth future trajectories of the notable objects and perform motion classification with heuristic threshold values on the trajectories as the ground-truth answers. We use L2 error as the evaluation metric.

\noindent\textbf{Q6. Prediction by Planning (Figure~\ref{fig:q6_illustration}):} If a notable object suddenly accelerates, decelerates, or changes directions, it may be difficult to predict its future trajectory by only observing its current and past locations and motions. However, if a notable object is also a CAV, it can directly share its planned future trajectory with other CAVs. That planning information could be used as a more accurate predicted future trajectory to help other CAVs' prediction and planning procedures. In Q6, we include the current planned future trajectories of other CAVs and the answer of Q4. Overall Notable Objects (Figure~\ref{fig:q4_illustration}) as the input context in the question. To generate the answer, we perform motion classification in the same manner as for Q5. We further identify whether other CAVs are notable objects or not in Q6's answer. To measure  performance, we calculate the binary classification accuracy in determining whether the model can correctly identify if other CAVs are notable objects or not.

\noindent\textbf{Q7. Overall Prediction (Figure~\ref{fig:q7_illustration}):} This question takes the answers from Q5. Prediction by Perception (Figure~\ref{fig:q5_illustration}) and Q6. Prediction by Planning (Figure~\ref{fig:q6_illustration}) as the input context and merges them to generate the final prediction answer. If the answer of Q6 indicates that another CAV is a notable object, the model is expected to learn to use that CAV's planned future trajectory as its predicted future trajectory. The merged prediction output is then used as the input context for subsequent planning tasks. To measure performance, we compute L2 errors on the prediction output.

\begin{figure*}[!t]
\centering
\includegraphics[width=0.98\textwidth]{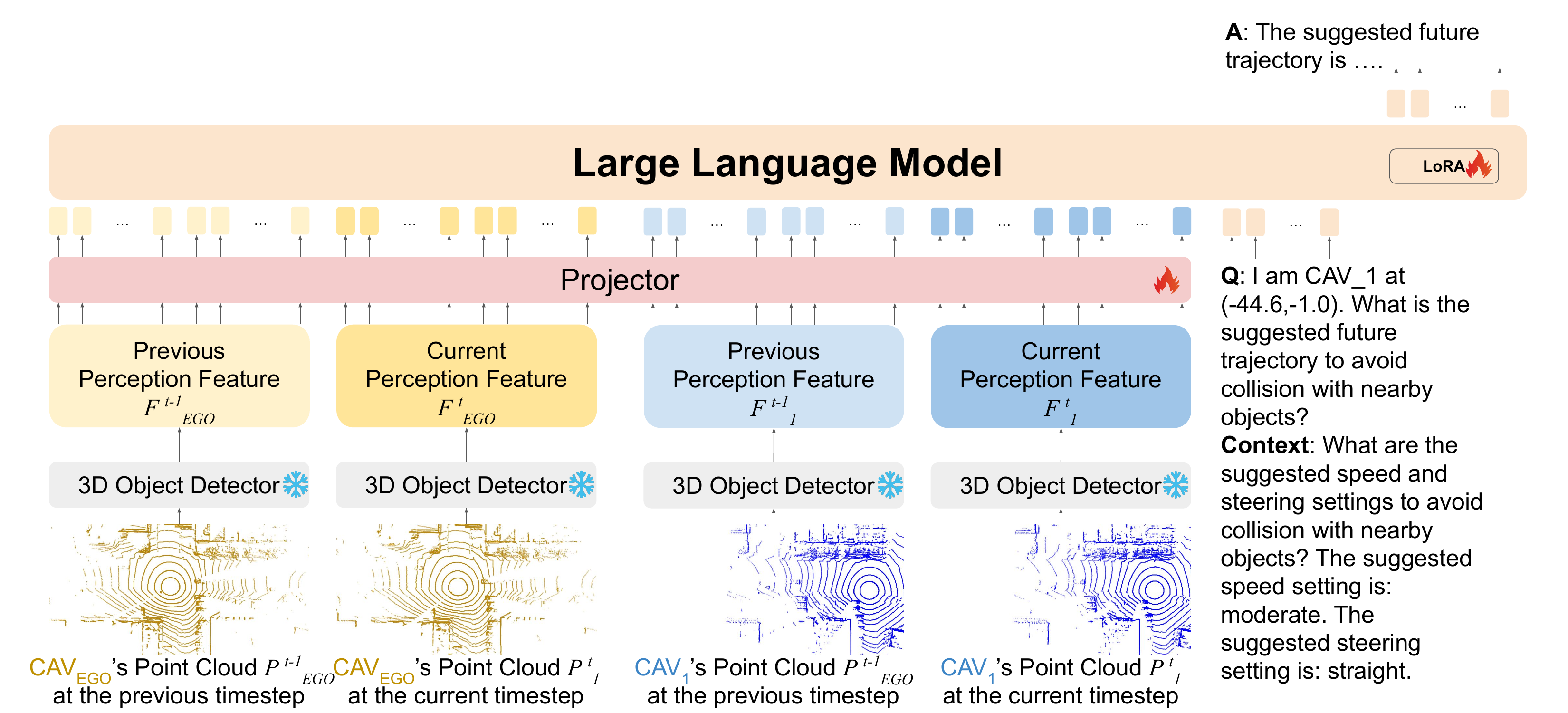}
\vspace{-5pt}
\caption[]
        {Model architecture of \namemethod.} 
        \vspace{-20pt}
        \label{fig:model}
\end{figure*}

\subsection{Planning QAs}
We first classify the suggested action's speed and steering settings, and then use them as the input context to provide the suggested future trajectory. 

\noindent\textbf{Q8. Suggested Action Classification (Figure~\ref{fig:q8_illustration}):} We use the prediction results from the answer of Q7. Overall Prediction (Figure~\ref{fig:q7_illustration}) as the input context and ask the MLLM to provide the suggested speed and steering settings. We adopt a similar approach from DriveLM~\cite{sima2023drivelm}. The speed setting can be one of the following 5 categories: \textit{fast}, \textit{moderate}, \textit{slow}, \textit{very slow}, and \textit{stop}. The steering setting can be one of the following 5 categories: \textit{left}, \textit{slightly left}, \textit{straight}, \textit{slightly right}, and \textit{right}. We classify the speed and steering settings by applying heuristic threshold values on the average difference between consecutive waypoint coordinates. To measure performance, we calculate the L1 errors between the output and ground-truth answer indices on the suggested speed and steering settings. For example, if the output speed and steering settings of a sample are \textit{fast} and \textit{slightly left}, while the ground-truth answer settings are \textit{very slow} and \textit{right}, the L1 error of this sample is $|0 - 3| + |1 - 4| = 6$.

\noindent\textbf{Q9. Suggested Trajectory (Figure~\ref{fig:q9_illustration}):} We use the suggested speed and steering setting from the answer of Q8. Suggested Action Classification (Figure~\ref{fig:q8_illustration}) as context and ask the LLM to provide the 6 waypoints of the suggested future trajectory for the next 3 seconds that avoids collisions. We use the ground-truth future trajectory as the ground-truth answer. To measure performance, we calculate the L2 errors and collision rates.


\section{\namemethod~Model}

\subsection{Architecture}

We build our \namemethod~model with LLaVA-v1.5-7b~\cite{liu2023llava} as the base MLLM architecture, as shown in Figure~\ref{fig:model}. Unlike the original LLaVA~\cite{liu2023llava} that uses an image encoder, we apply a LiDAR-based 3D object detector, PointPillars~\cite{lang2019pointpillar}, to extract perception features from each individual CAV's point cloud. 
Unlike \nameprior~\cite{chiu2025v2vllm}'s model that only uses the perception features at the current timestep, our model uses the perception features at the current and previous timesteps from all CAVs as the input of the project layers in the MLLM to generate the visual tokens. The MLLM takes the visual tokens and the language tokens from the question and the context as input and generates the final answer in the natural language format.

\subsection{Training and Inference}
We follow the similar training settings and hyperparameters from \nameprior~\cite{chiu2025v2vllm} and LLaVA~\cite{liu2023llava}. During training, we only train the projector layers and the LoRA~\cite{hu2022lora} parts of the model and freeze the remaining parts. Our model is trained with NVIDIA H100-80GB GPUs and a batch size of 32. We use Adam optimizer with a starting learning rate $2e^{-5}$ and train our model for 10 epochs on our \namedataset~training dataset. During testing inference, we follow our proposed graph-of-thoughts illustrated in Figure~\ref{fig:fig1} and \ref{fig:full_graph}. If two QA nodes are connected by a directed edge in the graph-of-thoughts, the inference model output of the parent node QA is used as the input context of the child node QA.

\section{Experiment}

\subsection{Baseline Methods}

We adopt \nameprior~\cite{chiu2025v2vllm} and its baseline methods with different fusion approaches, such as \textit{no fusion}, \textit{early fusion}, and \textit{intermediate fusion}, to compare with our proposed \namemethod. In \textit{no fusion}, the MLLM only takes a single CAV's perception features as visual input and answers the driving-related question. This baseline shows the performance in driving scenarios without cooperative perception. In \textit{early fusion}, the union of the point cloud of both CAVs is used to extract perception features, which are then used as the visual input of the MLLM. This approach usually requires high communication costs and is thus impractical for actual deployment. In \textit{intermediate fusion}, such as CoBEVT~\cite{xu2022cobevt}, V2X-ViT~\cite{xu2022v2xvit}, and AttFuse~\cite{xu2022opencood}, cooperative 3D object detectors are used to extract perception features as MLLM's visual input. These approaches usually achieve a good balance of performance and communication cost. \nameprior~\cite{chiu2025v2vllm} was proposed as a new fusion method that uses MLLM to fuse the scene-level and object-level features from multiple CAVs and achieves the best performance in V2V-QA~\cite{chiu2025v2vllm}'s perception and planning tasks.

\begin{table*}[!t]
\caption{
Testing performance of \namemethod~in the planning task of \namedataset~dataset, in comparison with baseline methods, which are adopted from V2V-LLM~\cite{chiu2025v2vllm}. To have a fair comparison to our proposed \namemethod, we modified baseline methods to also take perception features at the current and the previous timesteps as visual input. L2: L2 distance error. CR: Collision rate. Comm: Communication cost. In each column, the \textbf{best} results are in boldface, and the \underline{second-best} results are in underline.
\vspace{-10pt}
}
\begin{center}
\begin{tabular}{l cccc cccc c}
  \hline
  \hline
  \multirow{2}{*}{Method} &
  \multicolumn{4}{c}{L2 (m) $\downarrow$} & \multicolumn{4}{c}{CR (\%) $\downarrow$} & \multirow{2}{*}{Comm(MB) $\downarrow$} \\

  \cmidrule(lr){2-5} \cmidrule(lr){6-9}
  & \multicolumn{1}{c}{1s} & \multicolumn{1}{c}{2s} & \multicolumn{1}{c}{3s} & \multicolumn{1}{c}{Average} & \multicolumn{1}{c}{1s} & \multicolumn{1}{c}{2s} & \multicolumn{1}{c}{3s} & \multicolumn{1}{c}{Average} \\
  
  \hline
  \hline
  \textit{No Fusion}         & 3.47 & 5.79 & 8.26 & 5.84 & 1.48 & 4.24 & 7.72 & 4.48 & \textbf{0} \\
  \textit{Early Fusion}      & 3.48 & 5.61 & 7.82 & 5.63 & 1.16 & 3.51 & 5.66 & 3.44 & 1.9208 \\
  \hline
  \scriptsize{\textit{Intermediate Fusion}} \\ 
  AttFuse~\cite{xu2022opencood}         & 3.65 & 6.21 & 8.75 & 6.20 & 1.19 & 4.41 & 6.38 & 3.99 & \underline{0.4008} \\
  V2X-ViT~\cite{xu2022v2xvit}           & 3.46 & 5.80 & 8.19 & 5.81 & 1.45 & 4.24 & 6.59 & 4.09 & \underline{0.4008} \\
  CoBEVT~\cite{xu2022cobevt}            & 3.38 & 5.42 & 7.46 & 5.42 & 1.31 & 4.41 & 5.75 & 3.82 & \underline{0.4008} \\
  \hline
  \scriptsize{\textit{LLM Fusion}} \\
  V2V-LLM~\cite{chiu2025v2vllm}         & \underline{2.90} & \underline{4.91} & \underline{6.98} & \underline{4.93} & \underline{0.75} & \underline{2.87} & \underline{4.93} & \underline{2.85} & 0.4068 \\
  \namemethod~(Ours)                    & \textbf{1.65} & \textbf{2.63} & \textbf{3.59} & \textbf{2.62} & \textbf{0.12} & \textbf{1.92} & \textbf{3.45} & \textbf{1.83} & 0.4068 \\
  \hline
\end{tabular}
\label{tab:v2vllm_result}
\end{center}
\vspace{-10pt}
\end{table*}

\begin{figure*}[!t]
        \centering
        \begin{subfigure}[t]{0.32\textwidth}
            \centering 
            \includegraphics[width=\textwidth]{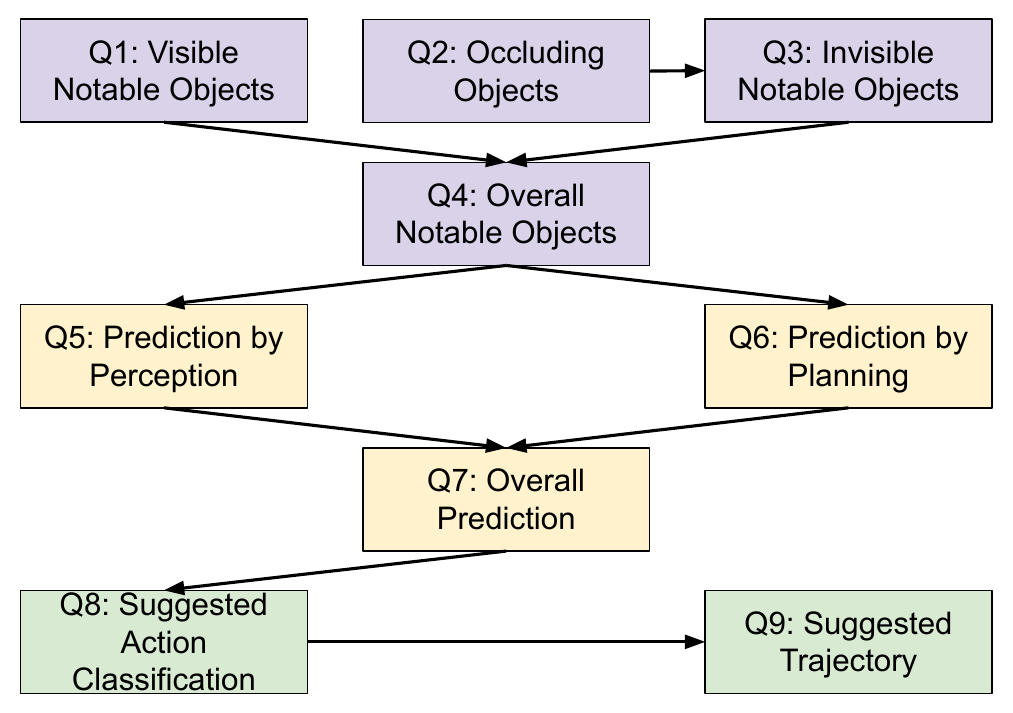}
            \caption[]%
            {{Our proposed graph-of-thoughts.}}    
            \label{fig:full_graph}
        \end{subfigure}
        \hfill
        \begin{subfigure}[t]{0.32\textwidth}  
            \centering 
            \includegraphics[width=\textwidth]{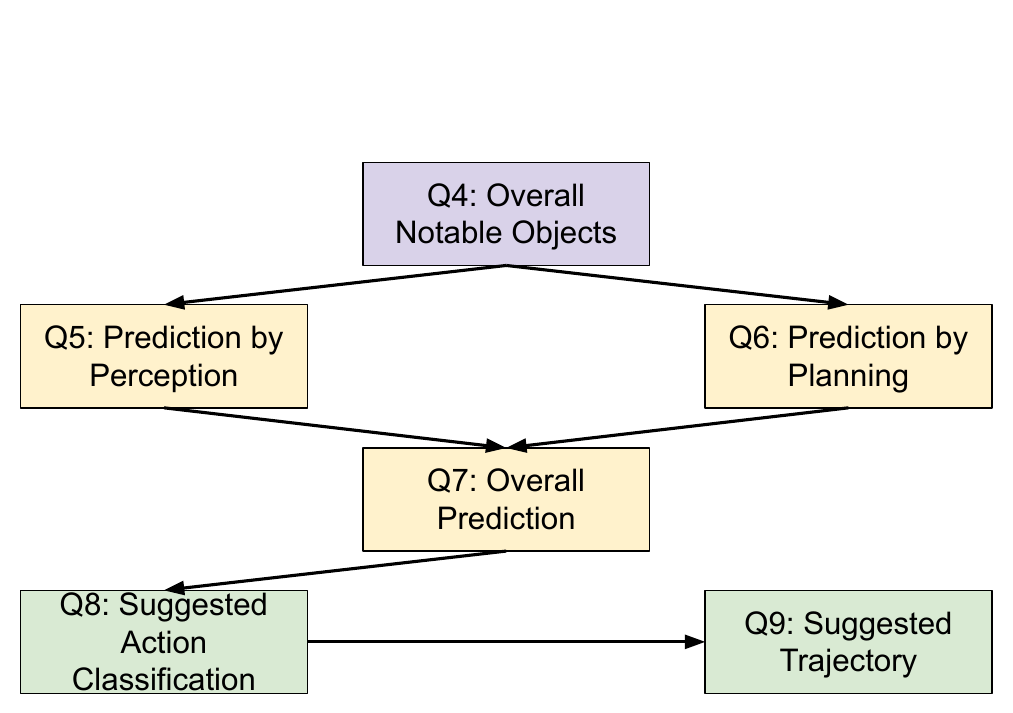}
            \caption[]%
            {{Simplified perception graph.}}    
            \label{fig:simplified_perception_graph}
        \end{subfigure}
        \hfill
        \begin{subfigure}[t]{0.32\textwidth}
            \centering 
            \includegraphics[width=\textwidth]{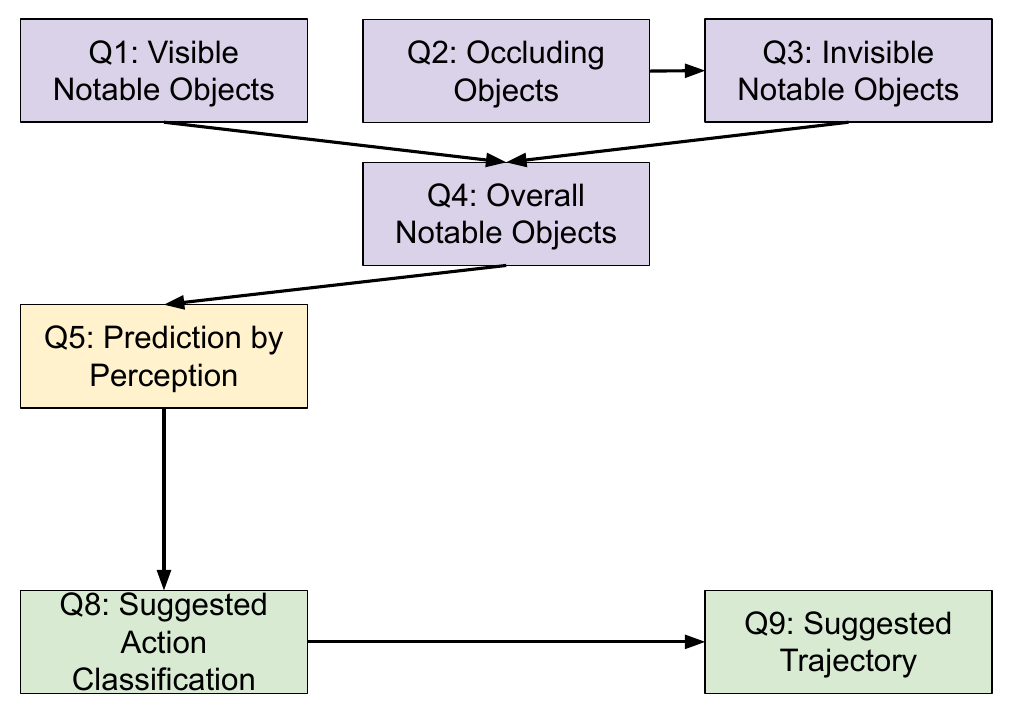}
            \caption[]%
            {{Simplified prediction graph.}}    
            \label{fig:simplified_prediction_graph}
        \end{subfigure}

        \vspace{-5pt}
        \caption[]
        {
        Different graph-of-thoughts structures for cooperative autonomous driving. The QA types include Perception (Q1 - Q4), Prediction (Q5 - Q7), and Planning (Q8 - Q9). If two nodes are connected by a directed edge, the answer of the parent node QA is used as the input context of the child node QA.
        } 
        \label{fig:all_gragh}
        \vspace{-5pt}
\end{figure*}   

\begin{table*}[t!]
\caption{
Testing performance of \namemethod~in all question-answering tasks of \namedataset~dataset, in comparison with different inference graphs. 
F1: F1 score. L2: L2 distance error. L1: L1 distance error. CR: Collision rate. Comm: Communication cost. In each column, the \textbf{best} results are in boldface.
\vspace{-10pt}
}
\begin{center}
\begin{tabular}{l cccc ccc c cc c}
  \hline
  \hline
  \multirow{2}{*}{Method} & Q1 & Q2 & Q3 & Q4 & Q5 & Q6 & Q7 & Q8 &
  \multicolumn{2}{c}{Q9} & \multirow{2}{*}{Comm(MB) $\downarrow$} \\

  \cmidrule(lr){2-2} \cmidrule(lr){3-3} \cmidrule(lr){4-4} \cmidrule(lr){5-5} \cmidrule(lr){6-6} \cmidrule(lr){7-7} \cmidrule(lr){8-8} \cmidrule(lr){9-9} \cmidrule(lr){10-11}

  & F1 $\uparrow$ & F1 $\uparrow$ & F1 $\uparrow$ & F1 $\uparrow$ & 
  L2 (m) $\downarrow$ & Accuracy & L2 (m) $\downarrow$ & L1 $\downarrow$ & L2 (m) $\downarrow$ & CR (\%) $\downarrow$ \\
  
  \hline
  \hline
 
  Simplified Perception                 & - & - & - & 58.0 & 8.18 & 82.0 & 7.66 & 0.0958 & 2.84 & 1.89 & 0.4068 \\
  Simplified Prediction                 & 52.5 & 30.1 & 44.0 & \textbf{60.8} & \textbf{8.05} & - & - & 0.0900 & 3.24 & 2.17 & 0.4068 \\
  \namemethod~(Ours)                    & 52.5 & 30.1 & 44.0 & \textbf{60.8} & \textbf{8.05} & \textbf{87.4} & \textbf{7.61} & \textbf{0.0876} & \textbf{2.62} & \textbf{1.83} & 0.4068 \\
  \hline
\end{tabular}
\label{tab:graph_result}
\end{center}
\vspace{-20pt}
\end{table*}

Our proposed \namedataset~dataset and the prior V2V-QA~\cite{chiu2025v2vllm} dataset use the same base dataset V2V4Real~\cite{xu2023v2v4real} and have the same final cooperative planning tasks: our Q9. Suggested Trajectory (Figure~\ref{fig:q9_illustration}) and V2V-QA~\cite{chiu2025v2vllm}'s Q5. Therefore, we can directly compare our method with the baseline methods on the final planning performance. Note that prior work \nameprior~\cite{chiu2025v2vllm}~only takes the perception features at the current timestep as the visual input. To have a fair comparison with our \namemethod, we modify their baseline models to also take the perception features at both of the current and previous timesteps as visual input, following our model architecture shown in Figure~\ref{fig:model}.

\subsection{Quantitative Results}
 
Table~\ref{tab:v2vllm_result} shows the testing performance of \namemethod~in the planning task of \namedataset~in comparison with baseline methods. 
Our newly proposed \namemethod~is seen to achieve the best final planning performance with the lowest L2 errors and collision rates compared to all baselines with different fusion approaches. In particular, the results indicate that the introduction of graph-of-thoughts reasoning designed for cooperative autonomous driving indeed boosts the performance in cooperative autonomous driving planning tasks.

\subsection{Ablation Study}
To further verify the impacts of our newly proposed ideas of \textbf{occlusion-aware perception} and \textbf{planning-aware prediction} in our graph-of-thoughts reasoning framework, as illustrated in Figures ~\ref{fig:fig1} and \ref{fig:full_graph}, we perform an ablation study on two additional graph structures: a \textit{simplified perception graph} shown in Figure~\ref{fig:simplified_perception_graph} and a \textit{simplified prediction graph} shown in Figure~\ref{fig:simplified_prediction_graph}. The ablation results are given in Table~\ref{tab:graph_result}.

\subsubsection{Simplified Perception Graph}
To verify the impact of \textbf{occlusion-aware perception}, we first create additional training data samples of Q4. Overall Notable Objects (Figure~\ref{fig:q4_illustration}) but without context. Then we train another model with newly generated data samples and the existing training dataset, and perform testing inference following the \textit{simplified perception graph} Figure~\ref{fig:simplified_perception_graph}. From Table~\ref{tab:graph_result}, we can see that the \textit{simplified perception graph} results in worse perception performance in Q4 and worse subsequent prediction and planning performance in Q7 and Q9, compared to our proposed \namemethod. 
This result indicates the importance of our occlusion-aware perception design in our final proposed graph-of-thoughts.

\subsubsection{Simplified Prediction Graph}
To verify the impact of \textbf{planning-aware prediction}, we use the same \namemethod~model but run testing inference on the \textit{simplified prediction graph}  (Figure~\ref{fig:simplified_prediction_graph}). From Table~\ref{tab:graph_result}, we can see that this \textit{simplified prediction graph} approach also results in poorer planning performance than our proposed \namemethod, due to the poorer prediction result, indicating the importance of our planning-aware prediction design. 
Furthermore, note that use of either simplified graph individually already offers  performance improvement over the baseline V2V-LLM~\cite{chiu2025v2vllm} (compare respective CR(\%) values in Tables~\ref{tab:v2vllm_result} and ~\ref{tab:graph_result}).


\section{Communication Cost}
The communication cost of \namemethod~is the same as the cost of the prior work \nameprior~\cite{chiu2025v2vllm}. 
Although \nameprior~\cite{chiu2025v2vllm} only uses the perception features at the current timestep and \namemethod~uses the features at the current and the previous timesteps, the same perception features only need to be transferred from a CAV to the MLLM once. The MLLM can save and reuse the perception features received in the current timestep for multiple QAs of the graph-of-thoughts in the current and the next timesteps. The texts of the intermediate questions and answers do not need to be transferred between CAVs and the MLLM unless requested. The MLLM can perform the inference with the graph-of-thoughts reasoning and send the final planning answer to the CAVs in the end. Therefore, the overall communication cost of \namemethod~is the same as prior work \nameprior~\cite{chiu2025v2vllm}.

\section{Qualitative Results}
\begin{figure}[!t]
\centering
\includegraphics[width=0.47\textwidth]{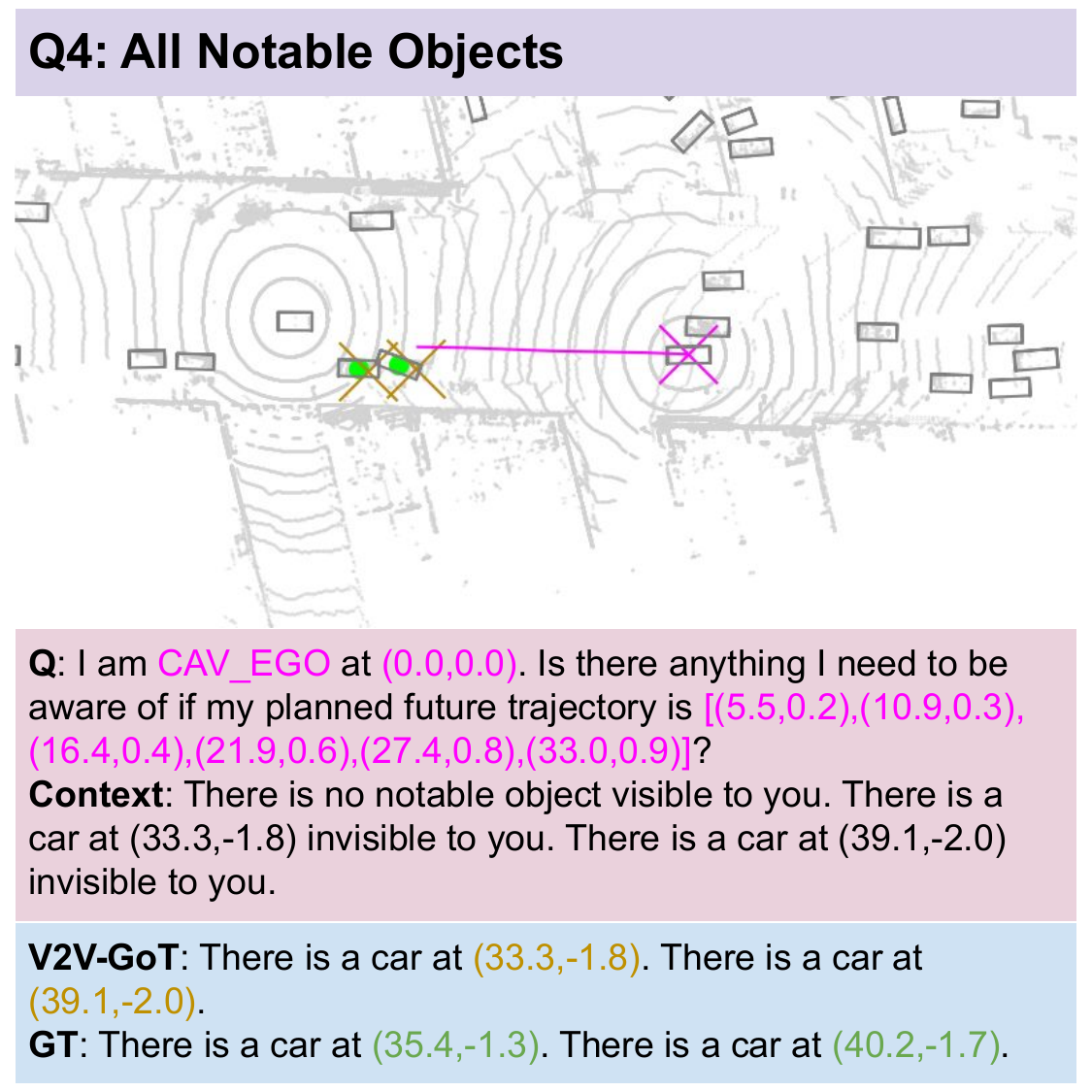}
\caption[]
        {Qualitative testing sample result of \namemethod~on Q4. Overall Notable Objects (Figure~\ref{fig:q4_illustration}). The context information is from the testing inference output of parent question Q3. Invisible Notable Objects (Figure~\ref{fig:q3_illustration}) and Q1. Visible Notable Objects (Figure~\ref{fig:q1_illustration}). \textcolor{magenta}{Magenta $\times$}: current location of the asking CAV. \textcolor{magenta}{Magenta curve}: reference trajectory in the question. \textcolor{olive}{Yellow $\times$}: model output. \textcolor{Green}{Green $\circ$}: ground-truth answer.
        } 
        \vspace{-20pt}
        \label{fig:q4_sample}
\end{figure}

\begin{figure}[!t]
\centering
\includegraphics[width=0.47\textwidth]{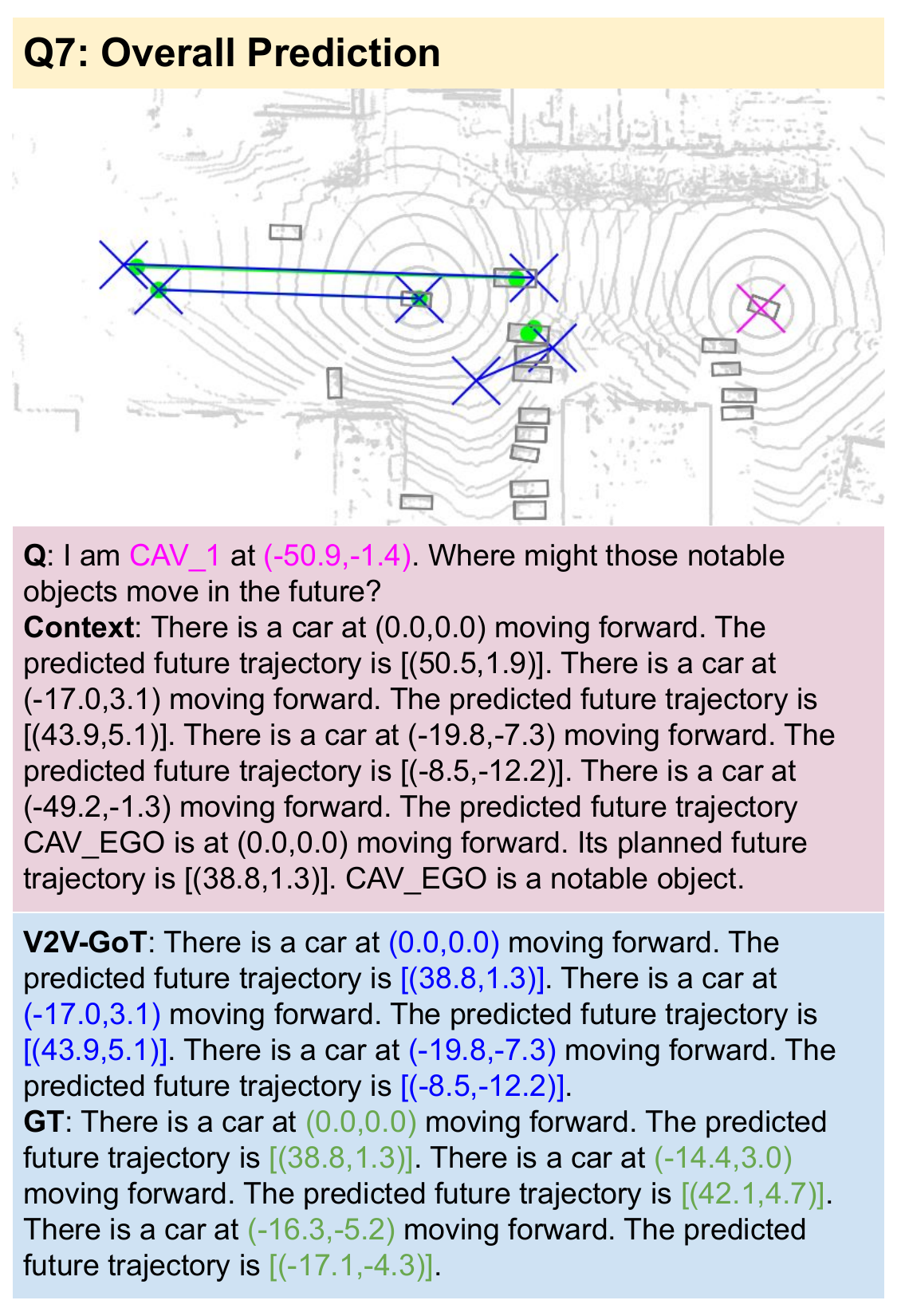}
\caption[]
        {Qualitative testing sample result of \namemethod~on Q7. Overall Prediction (Figure~\ref{fig:q7_illustration}). The context information is from the testing inference output of parent question Q5. Prediction by Perception (Figure~\ref{fig:q5_illustration}) and Q6. Prediction by Planning (Figure~\ref{fig:q6_illustration}). \textcolor{magenta}{Magenta $\times$}: current location of the asking CAV. \textcolor{blue}{Blue line and $\times$}: model output of the predicted future trajectories, starting, and ending waypoints of notable objects. \textcolor{Green}{Green line and $\circ$}: ground-truth answer.} 
        \vspace{-20pt}
        \label{fig:q7_sample}
\end{figure}

\begin{figure}[!t]
\centering
\includegraphics[width=0.47\textwidth]{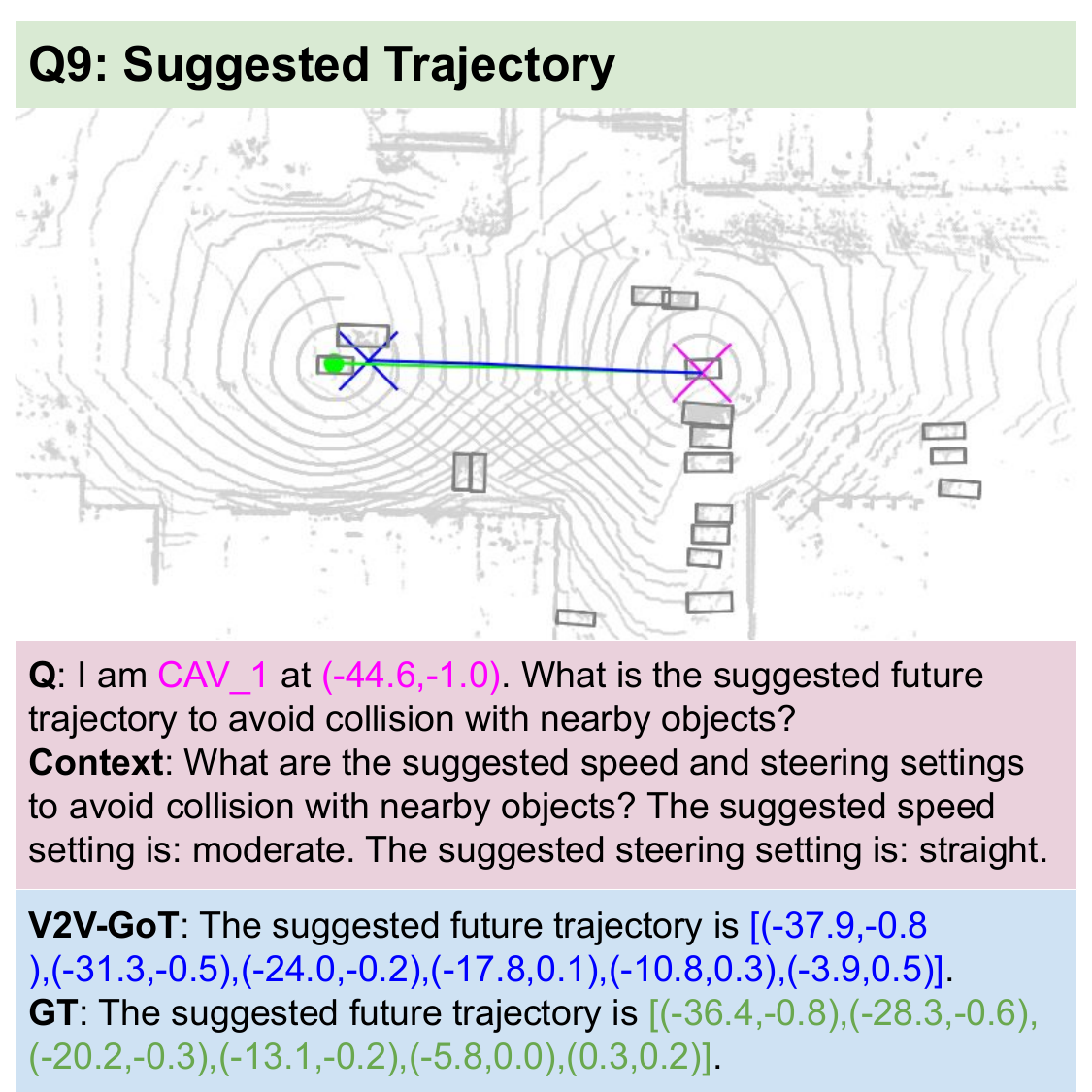}
\caption[]
        {Qualitative testing sample result of \namemethod~on Q9. Suggested Trajectory (Figure~\ref{fig:q9_illustration}). The context information is from the testing inference output of parent question Q8. Suggested Action Classification (Figure~\ref{fig:q8_illustration}). \textcolor{magenta}{Magenta $\times$}: current location of the asking CAV. \textcolor{blue}{Blue curve and $\times$}: model output of the suggested future trajectory and the ending waypoint. \textcolor{Green}{Green curve and $\circ$}: ground-truth answer.} 
        \vspace{-20pt}
        \label{fig:q9_sample}
\end{figure}

Figures~\ref{fig:q4_sample}, \ref{fig:q7_sample}, and~\ref{fig:q9_sample} show the qualitative testing results of our proposed \namemethod~in Q4. Overall Notable Objects (Figure~\ref{fig:q4_illustration}), Q7. Overall Prediction (Figure~\ref{fig:q7_illustration}), and Q9. Suggested Trajectory (Figure~\ref{fig:q9_illustration}), respectively. The input context information of each question comes from the model inference output of the associated parent question(s). In general, we can observe that our method generates perception, prediction, and planning output results close to ground-truth answers. More qualitative results can be found in the accompanying video.

\section{Conclusion}
In this work, we propose a novel graph-of-thoughts reasoning framework for MLLM-based cooperative autonomous driving. Our proposed graph-of-thoughts includes occlusion-aware perception and planning-aware prediction, which are designed to take advantage of V2V information sharing in cooperative driving scenarios and the multimodal understanding ability of MLLMs. To verify the effectiveness of our proposed ideas, we curate the \namedataset~dataset and develop the \namemethod~model. Our experimental results show that our proposed method outperforms all other baseline methods. Moreover, the ablation study further indicates that our innovative designs in the graph-of-thoughts improve overall cooperative perception, prediction, and planning performance. To facilitate open-source research, we have publicly released our \namedataset~dataset and \namemethod~code.


{\small
\bibliographystyle{IEEEtran}
\bibliography{egbib}
}



\end{document}